\documentclass{osa-article}

\journal{boe}
\articletype{Research Article}
\usepackage{enumitem}

\usepackage{float}
\usepackage[noend]{algpseudocode}
\usepackage{algorithmicx,algorithm}
\usepackage{graphicx}
\usepackage{multirow}

\newcommand{\mbf}[1]{\mathbf{#1}}

\begin{document}

\title{Annotation-efficient learning for OCT segmentation}

\author{Haoran Zhang\authormark{1}, Jianlong Yang\authormark{1,*}, Ce Zheng\authormark{2}, Shiqing Zhao\authormark{1}, and Aili Zhang\authormark{1}}

\address{\authormark{1}School of Biomedical Engineering, Shanghai Jiao Tong University, Shanghai, China\\
\authormark{2}Department of Ophthalmology, Xinhua Hospital Affiliated to Shanghai Jiao Tong University School of Medicine, Shanghai, China\\
\authormark{*}Corresponding Author: Jianlong Yang (jyangoptics@gmail.com)} %% email address is required



\begin{abstract}
Deep learning has been successfully applied to OCT segmentation. However, for data from different manufacturers and imaging protocols, and for different regions of interest (ROIs), it requires laborious and time-consuming data annotation and training, which is undesirable in many scenarios, such as surgical navigation and multi-center clinical trials. Here we propose an annotation-efficient learning method for OCT segmentation that could significantly reduce annotation costs. Leveraging self-supervised generative learning, we train a Transformer-based model to learn the OCT imagery. Then we connect the trained Transformer-based encoder to a CNN-based decoder, to learn the dense pixel-wise prediction in OCT segmentation. These training phases use open-access data and thus incur no annotation costs, and the pre-trained model can be adapted to different data and ROIs without re-training. Based on the greedy approximation for the k-center problem, we also introduce an algorithm for the selective annotation of the target data. We verified our method on publicly-available and private OCT datasets. Compared to the widely-used U-Net model with 100\% training data, our method only requires $\sim10\%$ of the data for achieving the same segmentation accuracy, and it speeds the training up to $\sim3.5$ times. Furthermore, our proposed method outperforms other potential strategies that could improve annotation efficiency. We think this emphasis on learning efficiency may help improve the intelligence and application penetration of OCT-based technologies. Our code and pre-trained model are publicly available at \url{https://github.com/SJTU-Intelligent-Optics-Lab/Annotation-efficient-learning-for-OCT-segmentation}.
\end{abstract}

%%%%%%%%%%%%%%%%%%%%%%%%%%  body  %%%%%%%%%%%%%%%%%%%%%%%%%%
\section{Introduction}
Among the \textit{in vivo} and \textit{in situ} tomographic imaging modalities of the human body, optical coherence tomography (OCT) has unique advantages in spatial resolution, which enables its tremendous success in clinical translation, such as ophthalmology, percutaneous and transluminal intervention, and dermatology \cite{huang1991optical,swanson2017ecosystem}. In the clinical practice of OCT, achieving quantitative biometrics through region-of-interest (ROI) segmentation, such as lesion and treatment area/volume, tissue layer thickness, and vessel density, is a prerequisite for standardized diagnostic and therapeutic procedures \cite{tian2016performance}. Due to the complexity and high throughput of clinical OCT data, tedious and time-consuming manual segmentation becomes a heavy burden for clinicians. In recent years, the rise of OCT-based angiography and medical robotics has further increased the need for automatic OCT segmentation \cite{kashani2017optical,draelos2021contactless}.\\
\indent The current methodological paradigm for automatic OCT segmentation is shifting from classical computer vision approaches (\textit{e.g.}, graph search/cut \cite{garvin2008intraretinal,chen2012three}, dynamic planning \cite{chiu2010automatic,kajic2012automated}, and active contouring \cite{yazdanpanah2010segmentation,gawlik2018active}) to deep learning-based approaches \cite{yanagihara2020methodological,litjens2019state}, which largely address the limitations of previous methods in dealing with fuzzy boundaries in diseased regions (\textit{e.g.}, fluid \cite{lu2019deep}, neovascularization \cite{wang2020automated}, edema \cite{hu2019automated}, drusen \cite{fang2017automatic}) and specific types of tissues (\textit{e.g.}, choroid-sclera interface \cite{zhang2020automatic}, capillaries \cite{ma2020rose}). Besides, benefiting from the end-to-end inference capabilities of deep learning, automatic segmentation can be synchronized with the OCT imaging process in real-time \cite{dos2019corneanet,borkovkina2020real}.\\
\indent However, achieving a trained deep neural network for OCT segmentation usually requires massive pixel-wise data annotation \cite{fang2021impact}, and the trained model only works well on data from the same manufacturers, using the same imaging protocols, and for the same ROIs. Several works have shown that the segmentation accuracy is severely degraded when the same subjects are captured with OCT devices from different manufacturers \cite{guan2021domain,chai2020perceptual,chai2021memory}. These problems bring nonnegligible inconvenience to applications in both research and clinical scenarios. For surgical navigation, ROI varies among different patients, organs, and lesions, posing enormous challenges for the generalization and real-time operation of deep learning-based segmentation tools. For multi-center clinical studies, different centers usually own OCT devices from different manufacturers, it is laborious to unify data from all sources and train a universal segmentation model. \\
\indent In this work, we propose an annotation-efficient learning method for OCT segmentation that significantly reduces annotation costs. We develop a progressive learning strategy by leveraging the self-supervised generative learning paradigm \cite{liu2021self,krishnan2022self}. We also introduce an algorithm for the selective annotation of the target data, which further contributes to the annotation efficiency. We describe our methods in Section 2. Their implementation and the datasets used in the experiments are detailed in Section 3. We give our experimental results in Section 4 and discuss the advantages and limitations of this work in Section 5. We draw our conclusion in Section 6.
\section{Methods}
\begin{figure}[!b]
\centering\includegraphics[width=13cm]{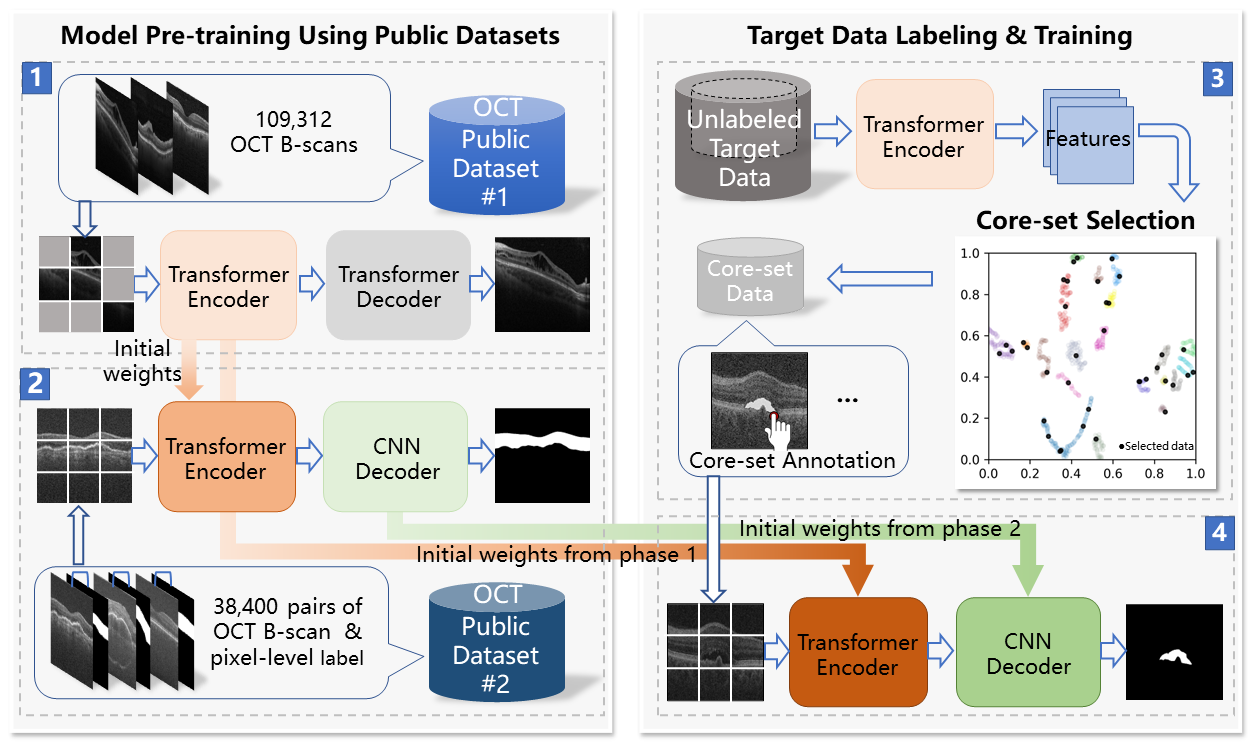}
\caption{Overview of our proposed method.}\label{fig:overview}
\end{figure}
Figure~\ref{fig:overview} is the overview of the proposed method. The pre-training strategy of our OCT segmentation model is inspired by the process of Human visual perception \cite{noton1971eye}, which follows a progressive manner by first learning the OCT imagery (Phase 1 in the upper left) and then the pixel-wise classification (Phase 2 in the bottom left). We use publicly-available large-scale datasets \cite{kermany2018identifying,farsiu2014quantitative} in these pre-training Phases thus inducing no annotation costs. We leverage the modeling capacity with self-attention of the Transformers \cite{vaswani2017attention,dosovitskiy2020image,khan2022Transformers}, and the scalability and generalization capacity of masked self-supervised generative learning \cite{he2022masked,wei2022masked}. For the target OCT data, we introduce an algorithm for selective annotation (Phase 3 in the upper right). It constructs a subset (core-set) that represents the entire target data in feature space. Only the B-scans in the core-set are annotated and used for fine-tuning the pre-trained model (Phase 4 in the bottom right). The initial weights of the Transformer encoder and the CNN decoder are inherited from the trained models in Phase 1 and Phase 2, respectively. Finally, the trained model can be used to infer the segmentation results of the remaining B-scans in the target OCT data.   
\subsection{Phase 1: Masked self-supervised generative pre-training}
\begin{figure}[!b]
\centering\includegraphics[width=13cm]{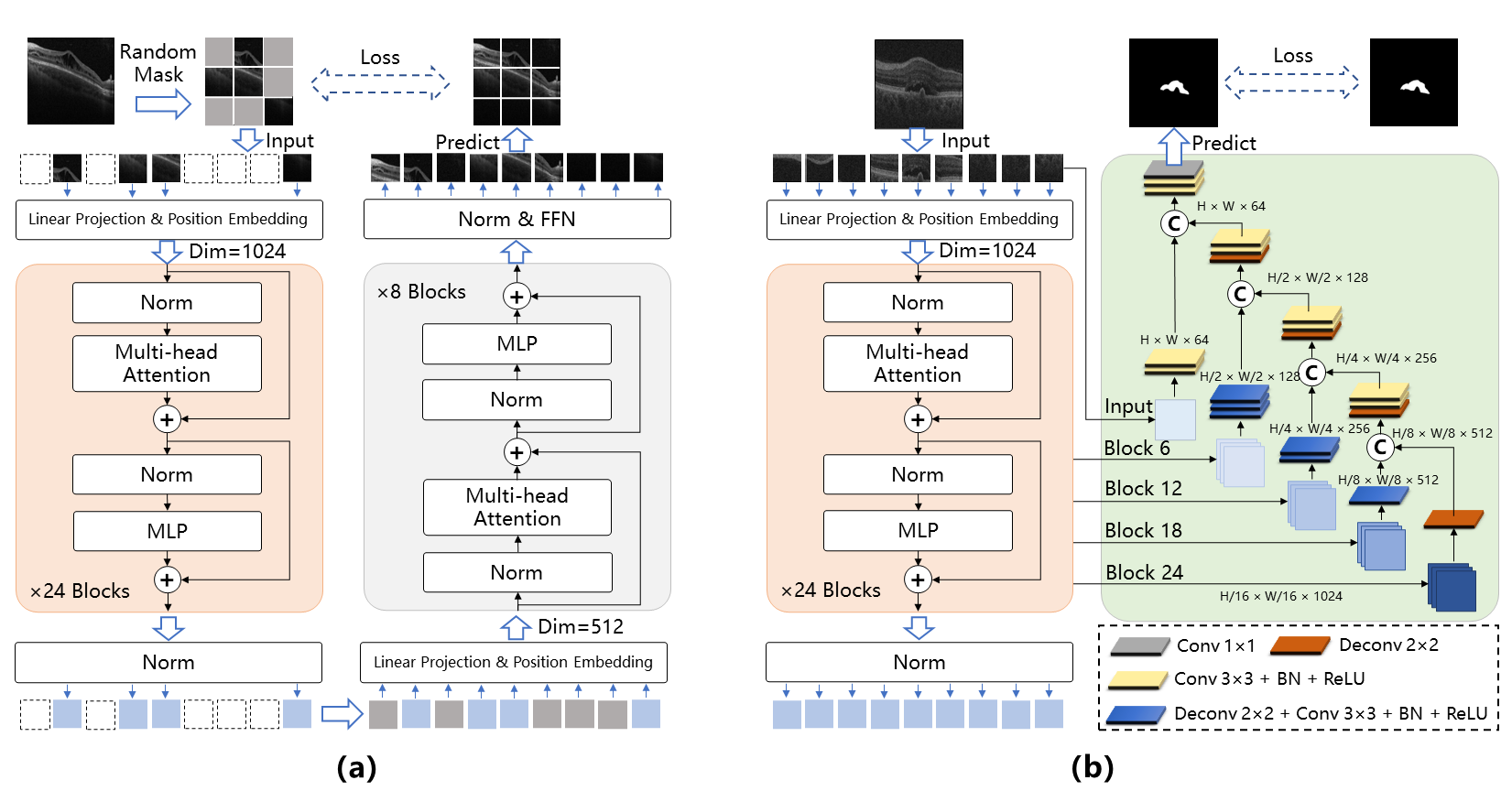}
\caption{Model architectures of (a) the masked self-supervised generative pre-training in Phase 1 and (b) the segmentation pre-training in Phase 2.}\label{fig:archs}
\end{figure}
We follow the learning strategy of the masked autoencoders (MAE) \cite{he2022masked}, in which the objective is to reconstruct missing pixels after randomly masking patches of input images. The detailed model architecture used in the masked self-supervised generative pre-training (Phase 1 in Fig.~\ref{fig:overview}) is illustrated in Fig.~\ref{fig:archs}(a). An OCT B-scan $\mbf{x} \in \mathbb{R}^{H \times W \times C}$ is divided into a sequence of non-overlapping patches $\mbf{x}_p \in \mathbb{R}^{N \times (P^2 \times C)}$, where $H$ and $W$ are the height and width of the original B-scan in pixels, respectively. $C$ is the number of channels. We set the patches to have the same height and width $P$. $N=HW/P^2$ is the resulting number of patches, which also serves as the effective input sequence length for the Transformer \cite{dosovitskiy2020image}. Then we randomly remove $M$ patches. The ratio between the $M$ and the number of all patches $N$ from an image is defined as the masking ratio. Only the remaining patches are flattened and mapped to $D$ dimensions with a trainable linear projection. We add position embeddings $\mbf{E}_{pos}$ to the patch embeddings to keep the positional information of the patches in the original OCT B-scan, which can be written as:
\begin{equation}
  \mbf{z}_0 = [\mbf{x}^1_p \mbf{E}; \, \mbf{x}^2_p \mbf{E}; \cdots; \, \mbf{x}^{N-M}_p \mbf{E} ] + \mbf{E}_{pos}, \;\;\;\;\; \mbf{E} \in \mathbb{R}^{(P^2 \times C) \times D},\, \mbf{E}_{pos}  \in \mathbb{R}^{(N-M) \times D}.
\end{equation}
\indent The embedding vectors are inputted into the Transformer encoder \cite{vaswani2017attention} that has a stack of blocks. Each block further consists of multi-head self-attention ($MSA$) and multi-layer perception ($MLP$) sub-blocks. We apply layer normalization ($LN$) before each sub-block and residual connections after each sub-block. These operations can be written as \cite{dosovitskiy2020image}:
\begin{align}
{{\mbf{z}}^\prime}_\ell &= MSA(LN({\mbf{z}_{\ell - 1}})) + {\mbf{z}_{\ell - 1}}, &\ell = 1\ldots L,\\
 % \mbf{z^\prime}_\ell &= \op{MSA}(\op{LN}(\mbf{z}_{\ell-1})) + \mbf{z}_{\ell-1}, & \ell=1\ldots L, \\
 {{\mbf{z}}}_\ell &= MLP(LN({{{\mbf{z}}^\prime}_{\ell}})) + {{{\mbf{z}}^\prime}_{\ell}}, &\ell = 1\ldots L,
  % \mbf{z}_\ell &= \op{MLP}(\op{LN}(\mbf{z^\prime}_{\ell})) + \mbf{z^\prime}_{\ell}, && \ell=1\ldots L,
\end{align}
where $l$ is the serial number of each sub-block. Then the encoded visible patches and mask tokens are inputted into the Transformer decoder, which adopts the architecture of the MAE decoder in \cite{he2022masked}. Each mask token is a shared, learned vector that indicates the presence of a missing patch to be predicted \cite{he2022masked,devlin2018bert}. The position embeddings are added to all tokens to keep their location information in the image. The output of the MAE decoder is reshaped to form a reconstructed image. The objective $\mathcal{L}_1$ is to minimize the mean squared error (MSE) between the reconstructed pixel value $\mathop y\limits^ \wedge$ and the corresponding pixel value $y$ of original masked patch in the normalized pixel space:
\begin{equation}
    \mathcal{L}_1=\frac{1}{N_1}\sum^{N_1}_{i=1}(\mathop {{y_i}}\limits^ \wedge   - {y_i})^2,
\end{equation}
where $N_1$ is the number of pixels of all masked patches in a training image.
\subsection{Phase 2: Segmentation pre-training}
We construct the segmentation pre-training model by connecting the trained Transformer encoder in Phase 1 and a CNN-based decoder, except that the input of the Transformer encoder is the sequence of OCT patches without any masking operation. Inspired by the architectures of the U-Net and its successors \cite{ronneberger2015u,hatamizadeh2022unetr}, we merge the features from multiple resolutions of the Transformer encoders with the CNN decoder via skip connection as illustrated in Fig.~\ref{fig:archs}(b). Specifically, we extract the sequenced features from the output of uniformly-spaced sub-blocks of the Transformer encoder and reshape the size of the sequenced feature $\frac{{HW}}{{{P^2}}} \times D$ into $\frac{H}{P} \times \frac{W}{P} \times D$ as feature maps. Four up-sampling steps are implemented and the number of channels is halved progressively. Each step consists of up-sampling the low-scale feature maps by deconvolution and concatenation in the channel dimension of feature maps. We process them with convolutional layers, each subjected to batch normalization (BN) and ReLU activation. The same steps are repeated until the size of two-dimensional feature maps equals the original input. The last feature maps are processed by convolution and Sigmoid activation to generate pixel-wise segmentation prediction. The objective of this training Phase $\mathcal{L}_2$ is to minimize the summation of a binary cross entropy (BCE) loss $\mathcal{L}_{BCE}$ and a S{\o}rensen–Dice coefficient (DICE) loss $\mathcal{L}_{DICE}$ between the prediction of the CNN decoder $p$ and the corresponding labeled segmentation ground-truth $g$:
\begin{equation}
\mathcal{L}_2=\mathcal{L}_{BCE}+\mathcal{L}_{DICE}=-\frac{1}{N_2}\sum^{N_2}_{i=1}[g_i\log p_i+(1-g_i)\log (1-p_i)]+1-\frac{2\sum_{i=1}^{N_2} p_i g_i}{\sum^{N_2}_{i=1}p_i+\sum^{N_2}_{i=1}g_i}, 
\end{equation}
where $N_2$ is the number of pixels in a training image. Note that, although we focus on the foreground-to-background (binary) segmentation of OCT in this paper, our method can adapt to multi-class segmentation tasks (\textit{e.g.}, retinal layers) by simply modifying the segmentation objective above. 

\subsection{Phase 3: Selective annotation}
Random selection is straightforward but may introduce redundancy and inefficiency, which exist among adjacent B-scans from the same case (each case refers to OCT acquisition once from a subject). Here we propose a core-set selection algorithm for improving the selection efficiency inspired by the methodologies in active learning and core-set selection \cite{sener2018active}. Our aim is to find a small subset $S$ (core-set) that could efficiently represent the given training set of the target data $U$ with a budget of $s$ samples (the budget refers to the number of the chosen B-scans for annotation). Specifically, we try to find $s$ samples to geometrically approximate the feature space of $U$ with minimal distances to other samples. Here we use the Euclidean distance as the measure in feature space:   
\begin{equation}
{D_{i,j}} = {\mathop{\rm D}\nolimits} (f({u_i}),f({u_j})) = \| f({u_i}) - f({u_j})\|,
\end{equation} 
where $u_i$ and $u_j$ are arbitrary two samples in $U$. $f$ refers to the operations including the feature map acquisition of a sample via the trained Transformer encoder from Phase 1 and the reshaping of the feature map into a feature vector. Then this sample selection problem can be seen as the k-center problem and solved by the greedy approximation \cite{liang2016simple}.\\
\indent The detailed solution is described in Algorithm \ref{algorithm}. We first successively select one representative sample for $S$ from each case in $U$ ($n$ cases in $U$ in total), based on the observation that the cases are usually from different subjects, thus having natural separations in feature space. These $n$ samples are selected via the minimization of the overall distances to other samples in the same case. The remaining $s-n$ samples for $S$ are selected from the remaining samples in $U$ successively. 
% For the $i$-th selected sample ($n<i \le s$), it is chosen via the minimization of its maximal distances to the already selected $i-1$ samples. 
For the $i$-th selected sample ($n<i \le s$), it is chosen to minimize the largest distance between a data in existing unselected dataset and its nearest data in already selected subset. 

\begin{figure}[!h]
\centering\includegraphics[width=11cm]{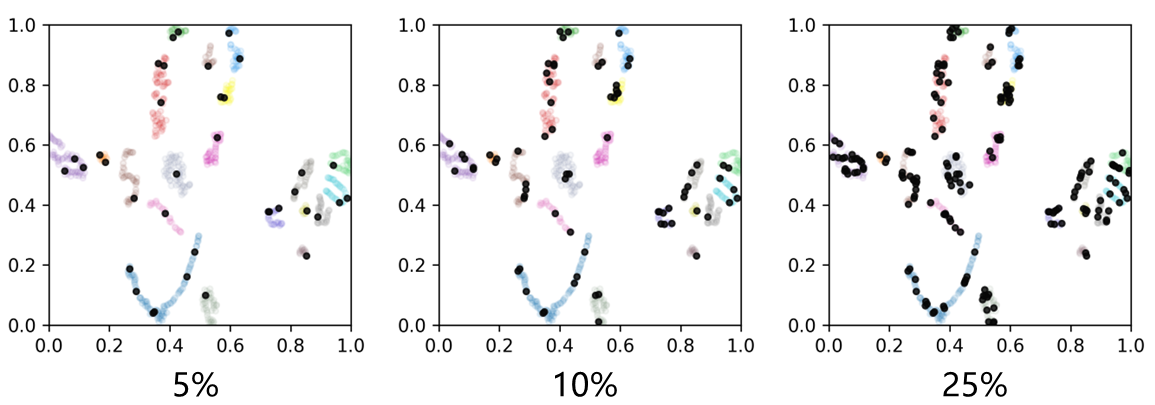}
\caption{t-SNE visualization of the performance of our selective annotation method with the data budgets of 5\%, 10\%, and 25\%. The black dots denote samples of the core-set data. The colored dots denote the entire unlabeled target data.}\label{tsne}
\end{figure}

We use the t-SNE method \cite{van2008visualizing} to visualize the performance of our selective annotation algorithm as shown in Fig~\ref{tsne}. For the data budgets of 5\%, 10\%, and 25\%, the selected samples (black dots) can sufficiently cover the feature space of a training set.
\begin{algorithm}[t!]
\caption{Selective annotation} \label{algorithm} %算法的名字
\hspace*{0.02in} {\bf Input:} %算法的输入， \hspace*{0.02in}用来控制位置，同时利用 \\ 进行换行
feature vectors of unlabeled dataset \textit{U} (including $n$ cases $U=\{ {N^i}\} _{i = 1}^n$, equalling to $m$ images $U=\{ {u_i}\} _{i = 1}^m $), data budget \textit{s} ($n < s <m$)\\
\hspace*{0.02in} {\bf Output:} %算法的结果输出
selected core-set \textit{S} ($S \subseteq U$)
\begin{algorithmic}[1]
\State  Initialize the selected core-set $\textit{S}=\{\}$  % \State 后写一般语句
\For{$i$ = 1, 2, ..., $n$} % For 语句，需要和EndFor
\State Initialize an empty distance vector \textit{D} with size (length(\textit{${N^i}$}))
\For {$j$ = 1, 2, ..., length(\textit{${N^i}$})}
\State ${D_j} \leftarrow \sum\nolimits_{a = 1}^{length({N^i})} {{\mathop{\rm D}\nolimits} (f(N_j^i),f(N_a^i))}$
\EndFor
\State $y \leftarrow \arg {\min _{x \in \{ 1, 2, ..., {\mathop{\rm length}\nolimits} ({N^i})\} }}{D_x}$
\State $U.{\mathop{\rm remove}\nolimits} (N_y^i)$ and $S.{\mathop{\rm append}\nolimits} (N_y^i)$
\EndFor
\For{$i$ = $n+1$, $n+2$,..., \textit{s}} % For 语句，需要和EndFor
%对应\State 

\State Initialize an empty distance matrix \textit{D} with size (length(\textit{U}), $i-1$)
\For {a = 1, 2, ..., length(\textit{U})}
\For {b = 1, 2, ..., $i-1$}
\State ${D_{a,b}} \leftarrow {\rm{D}}(f({u_a}),f({S_b}))$ 
\EndFor
\EndFor
\State $z \leftarrow \arg {\max _{x \in \{ 1,2,...,{\mathop{\rm length}\nolimits} (U)\} }}{\min _{y \in \{ 1,2,...{\mathop{\rm length}\nolimits} (S)\} }}({D_{x,y}})$
\State $U.{\mathop{\rm remove}\nolimits} ({u_z})$ and $S.{\mathop{\rm append}\nolimits} ({u_z})$
\EndFor
\State \Return selected core-set \textit{S}
\end{algorithmic}
\end{algorithm}
\section{Experimental settings}
\subsection{Datasets}
We use open-access (OA) datasets in the pre-training Phases of our model, which induces no annotation costs. In Phase 1, we use an OA dataset that contains 109,312 OCT B-scans taken from 4,686 patients using Heidelberg Spectralis OCT system \cite{kermany2018identifying} (hereinafter referred to as the OCT2017 dataset). These B-scans have four classes of image-level labels including choroidal neovascularization, diabetic macular edema, drusen, and normal. Because Phase 1 is based on self-supervised generative learning, we do not use these labels in the training. In Phase 2, we employ a dataset that contains 38,400 OCT B-scans with pixel-level labels from 269 age-related macular degeneration patients and 115 normal subjects \cite{farsiu2014quantitative} (hereinafter referred to as the DukeOCT dataset). These data were collected on a Bioptigen OCT system. They have manual segmentation boundaries of the inner limiting
membrane (ILM), the inner aspect of retinal pigment epithelium drusen complex (RPEDC), and the outer aspect of Bruch’s membrane (BM). We use the total retinal region (covered by ILM and BM) as the foreground label in the training of Phase 2.\\
\indent We evaluate our method using three types of ROIs including the subretinal fluid (SRF), pigment epithelial detachment (PED), and choroid, from an OA OCT dataset named RETOUCH \cite{bogunovic2019retouch} and a private OCT dataset. The RETOUCH dataset was acquired with devices from different manufacturers including ZEISS, Heidelberg, and Topcon \cite{bogunovic2019retouch}. The private dataset was collected on a homemade SD OCT system (hereinafter referred to as the CHOROID dataset). All the data are randomly split into training and testing sets on a patient basis. For the SRF, we use 969 B-scans from 23 cases for training and 471 B-scans from 14 cases for testing. For the PED, we use 737 B-scans from 19 cases for training and 701 B-scans from 14 cases for testing. For the choroid, we uses 220 B-scans from 11 cases for training and 100 B-scans from 5 cases for testing. Because these datasets are relatively small, We also verified the results below using the K-fold cross-validation (Please see the Supplemental Document for the description).
\\

\subsection{Implementations} 
We implement our code using PyTorch\footnote{http://pytorch.org/} and trained it on a personal computer with an Nvidia 3090 GPU (24G onboard memory). all the OCT B-scans are resized to $224 (H) \times 224 (W) \times 3 (C)$ pixels. Although the original OCT data is collected in gray-scale ($C=1$), we use the default RGB channels used in the vision Transformer models \cite{dosovitskiy2020image,he2022masked} for simplicity. In Phase 1, we employ the ViT-Large \cite{dosovitskiy2020image} as the Transformer encoder with the stack of 24 consecutive sub-blocks and embedding size of $D = 1024$ as shown in Fig.~\ref{fig:archs}. The patch size sets $P \times P = 16 \times 16$. we set the masking ratio $M/N$ of $75\%$. The settings of the Transformer decoder follow the lightweight design used in \cite{he2022masked} with an embedding size $D$ of 512 and 8 consecutive sub-blocks. We employ the weights trained on ImageNet \cite{he2022masked} as initialization and used the OCT2017 dataset \cite{kermany2018identifying} to fine-tune the model for 300 epochs.\\
\indent In Phase 2, we extract the sequenced features from the output of uniformly-spaced (${6^{{\rm{th}}}}$, ${12^{{\rm{th}}}}$, ${18^{{\rm{th}}}}$, ${24^{{\rm{th}}}}$) sub-blocks of the Transformer encoder. Four up-sampling steps are implemented and the number of channels is halved progressively, \textit{i.e.}, from 512 to 64. Each step consists of up-sampling the low-scale feature maps by $2 \times 2$ deconvolution and concatenation in the channel dimension of feature maps. At the bottleneck, \textit{i.e.}, the output of the last sub-block (${24^{{\rm{th}}}}$), we up-sample the transformed feature maps by convolution operation enlarging $2 \times $ resolution. And then we concatenate the enlarged feature maps with the same scale feature maps from ${18^{{\rm{th}}}}$ sub-block. We process them with two $3 \times 3$ convolutional layers. We use the trained weights of the Transformer encoder in Phase 1 for training the CNN decoder. The DukeOCT dataset \cite{farsiu2014quantitative} is used for training 100 epochs in this Phase. We use the AdamW \cite{loshchilov2018decoupled} optimizer with an initial learning rate of \textit{1e-4}. The learning rate is scheduled as cosine decay \cite{loshchilov2016sgdr} and combines with a warm-up period \cite{goyal2017accurate} of 10 epochs.\\
\indent In Phase 4, we employ the weights of the trained Transformer encoder in Phase 1 and the trained CNN decoder in Phase 2 as initialization. The dataset used for training is the core-set constructed in Phase 3. We set a batch size of 4. We use the Adam \cite{kingma2014adam} optimizer (${\beta _1} = 0.9$, ${\beta _2} = 0.999$) with an initial learning rate of \textit{0.5e-4}. We adopt a dynamically varying learning rate scheduler. When the metric stops improving (no improvement for 5 consecutive times), the scheduler decreases the learning rate by $10\%$. To fairly compare the training times under different conditions, we used the EarlyStopping strategy, where the training process is stopped until the validation loss does not progress for 10 consecutive epochs. To keep the semantics of OCT images, we only employ random horizontal flipping for data augmentation. The Dice similarity coefficient (DSC) score is employed to evaluate the segmentation results, which the definition is as follows:
\begin{equation}
DSC = 2 \times \frac{{\left| {P \cap G} \right|}}{{\left| P \right| + \left| G \right|}}
\end{equation}
where $P$ and $G$ denote the target area of our segmentation results and ground-truth, respectively. The DSC varies from 0 (non-overlapping) to 1 (entire-overlapping).

%%%%%%%%%%%%%%%%%%%%%%%%%%%%%results%%%%%%%%%%%%%%%%%%%%%%%%%%%%%%%%%%%%%%%%
\section{Results}
\subsection{Comparison with U-Net}
\begin{figure}[!h]
\centering\includegraphics[width=13cm]{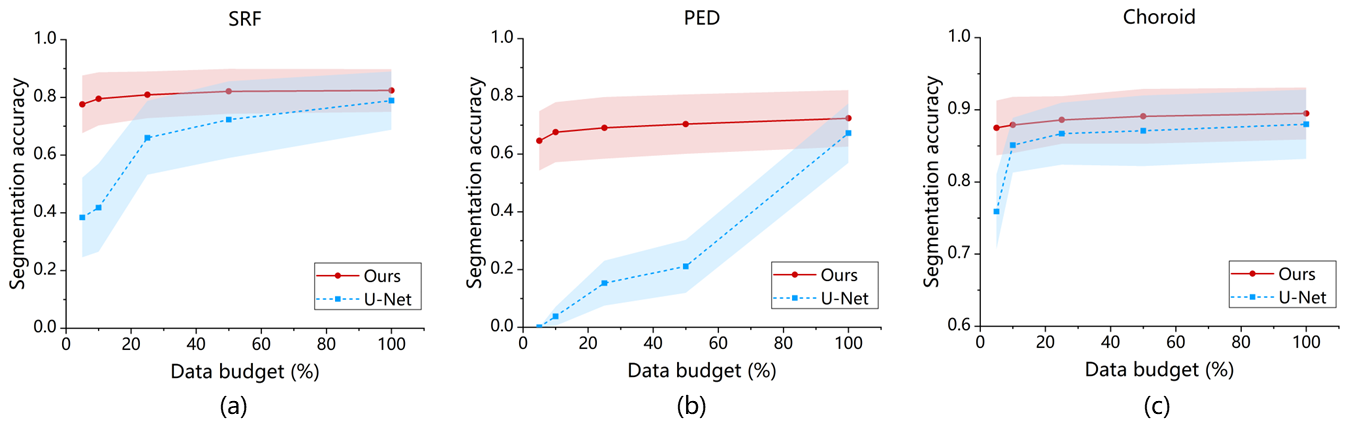}
\caption{Comparison of segmentation accuracy (DSC) with the U-Net results under different data budgets (5\%, 10\%, 25\%, 50\%, and 100\%) for the ROIs (a) SRF, (b) PED, and (c) Choroid.}\label{Figure 4}
\end{figure}
\begin{figure}[!h]
\centering\includegraphics[width=12cm]{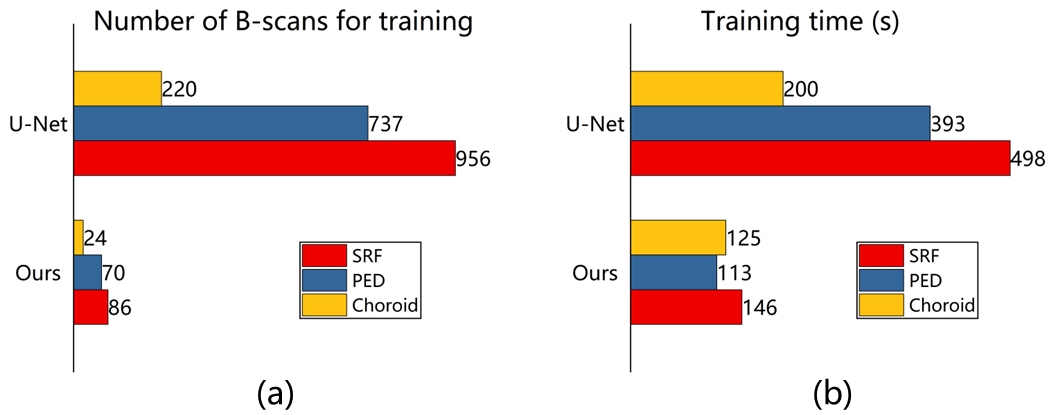}
\caption{The data (a) and training time (b) costs of our method for achieving the same segmentation accuracy using the U-Net with 100\% training data.}\label{Figure 5}
\end{figure}
\begin{figure}[!h]
\centering\includegraphics[width=13cm]{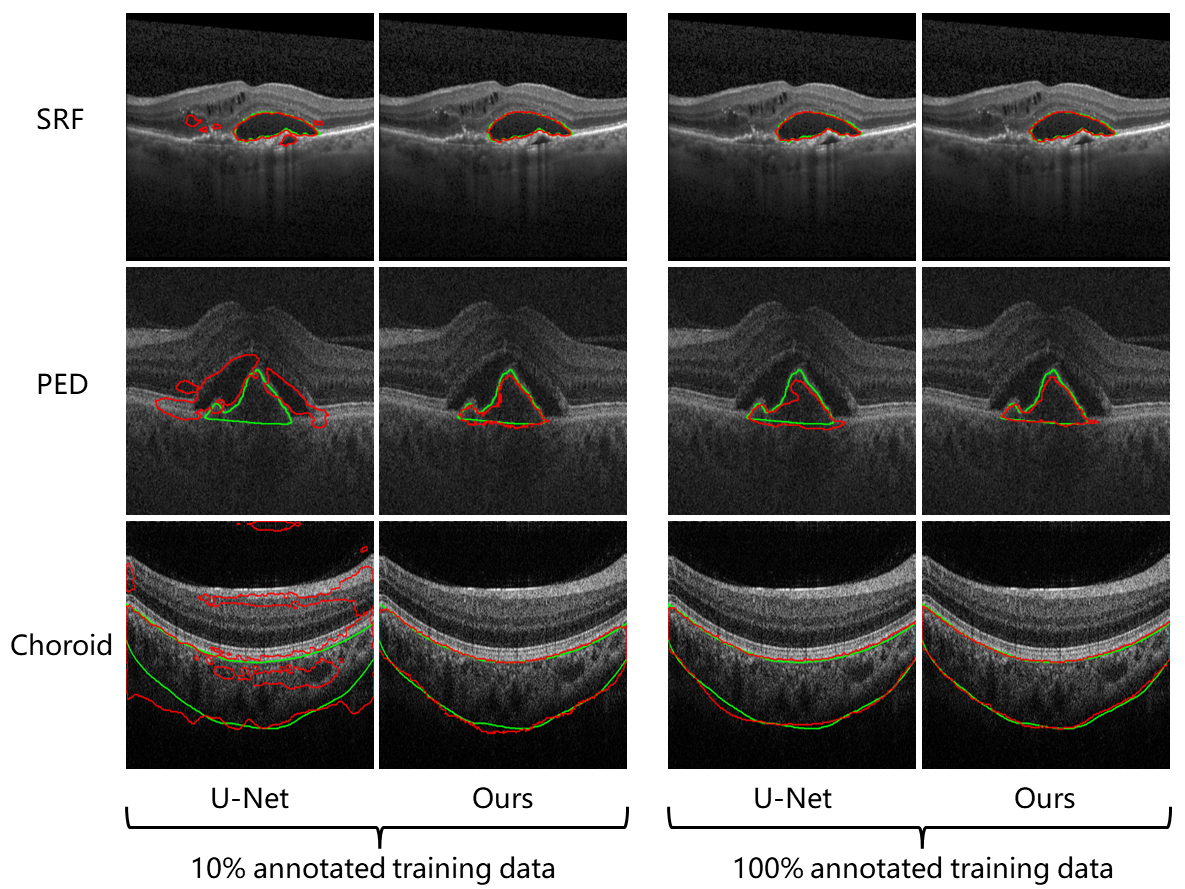}
\caption{Qualitative comparison of our method with the U-Net under the data budgets of 10\% (left) and 100\% (right). The ground truth is labeled in green.}\label{Figure 6}
\end{figure}
We use our model and the U-Net to segment the three types of ROIs with different data budgets ranging from 5\% to 100\%. The results of the SRF, PED, and Choroid are shown in Fig.~\ref{Figure 4}(a), (b), and (c), respectively. Inside the sub-figures, the dots connected with solid and dashed lines indicate mean values and the translucent areas illustrate standard deviations. We use red and blue colors to label the results of our method and the U-Net, respectively. Generally, for the U-Net, the segmentation performance improves as the data budget increases. While our method can achieve high segmentation accuracy even with 5\%-10\% of training data. It is worth noting that the morphological features of the Choroid do not vary much in different locations of the fundus and for different cases \cite{zhang2020automatic}, so the U-Net can also obtain a high segmentation performance with less training data.\\
\indent For the segmentation accuracy of the U-Net with 100\% training data, we plot the specific number of B-scans used in the training of our model performing approximative accuracy in Fig.~\ref{Figure 5}(a). We achieve savings in training data of 91.0\%, 90.5\%, and 89.1\% for the ROIs SRF, PED, and Choroid, respectively. Because less data is used in training, we also achieve improvements in training speed by up to $\sim3.5$ times as shown in Fig.~\ref{Figure 5}(b). It should be noted that only the re-training time using the selected target data (Phase 4) is compared, because our pre-trainings in Phases 1 and 2 were done once and for all.\\
\indent Figure~\ref{Figure 6} demonstrates the examples of the automatic segmentation results using our proposed method and the U-Net with 10\% (left) and 100\% (right) of the training data for the three types of ROIs. The red and green ones are the results and ground truths respectively. We can see a significant improvement in accuracy using our method under a data budget of 10\%.
\subsection{Ablation studies} \label{Ablation studies}
To evaluate the effectiveness of training Phase 1 and 2 designed in our proposed method, we perform ablation studies and the results are shown in Table \ref{tab:my-table1}. We compare the results under the three ROI types and the data budgets of 10\% and 100\%.\\
\begin{table}[!h]
\centering
\caption{Ablation studies of Phase 1 and 2. The data format in the table is the mean (standard deviation).}
\label{tab:my-table1}
\resizebox{\textwidth}{!}{%
\begin{tabular}{cc|cc|cc|cc}
\hline
\multicolumn{2}{c|}{\multirow{2}{*}{Ablation settings}} & \multicolumn{2}{c|}{SRF} & \multicolumn{2}{c|}{PED} & \multicolumn{2}{c}{Choroid} \\ \cline{3-8} 
\multicolumn{2}{c|}{} & 10\% & 100\% & 10\% & 100\% & 10\% & 100\% \\ \hline
\multicolumn{1}{c|}{W/o Phase 1 \& Phase 2} & N/A & 0.514 (0.268) & 0.555 (0.297) & 0.196 (0.190) & 0.283 (0.164) & 0.845 ()0.066 & 0.873 (0.073) \\ \hline
\multicolumn{1}{c|}{\multirow{3}{*}{W/ Phase 1 \& w/o Phase 2}} & ImageNet & 0.758 (0.232) & 0.802 (0.187) & 0.643 (0.200) & 0.689 (0.206) & 0.873 (0.075) & 0.894 (0.078) \\ \cline{2-8} 
\multicolumn{1}{c|}{} & OCT2017 & 0.716 (0.239) & 0.777 (0.189) & 0.414 (0.214) & 0.633 (0.204) & \textbf{0.880 (0.082)} & 0.891 (0.085) \\ \cline{2-8} 
\multicolumn{1}{c|}{} & \begin{tabular}[c]{@{}c@{}}ImageNet + \\ OCT2017\end{tabular} & 0.773 (0.212) & 0.811 (0.180) & 0.661 (0.217) & 0.715 (0.196) & 0.875 (0.072) & \textbf{0.895 (0.079)} \\ \hline
\multicolumn{1}{c|}{W/ Phase 1 \& Phase 2} & \begin{tabular}[c]{@{}c@{}}ImageNet + \\ OCT2017\end{tabular} & \textbf{0.795 (0.186)} & \textbf{0.824 (0.148)} & \textbf{0.676 (0.209)} & \textbf{0.724 (0.196)} & 0.879 (0.077) & \textbf{0.895 (0.073)} \\ \hline
\end{tabular}%
}
\end{table}
\begin{table}[!h]
\centering
\caption{Comparison of the proposed selection with random selection and uniform selection. The data format in the table is the mean (standard deviation).}
\label{tab:my-table2}
\resizebox{12cm}{!}{%
\begin{tabular}{cc|ccccc}
\hline
\multicolumn{2}{c|}{Setting}                                                           & 5\%                               & 10\%                              & 25\%                              & 50\%                              & 100\%                  \\ \hline
\multicolumn{1}{c|}{\multirow{3}{*}{SRF}}     & Ours                     & \textbf{0.776 (0.199)}            & \textbf{0.795 (0.186)}            & \textbf{0.809 (0.161)}            & \textbf{0.821 (0.155)}            & \textbf{0.824 (0.148)} \\
\multicolumn{1}{c|}{}                         & Random selection                       & 0.767 (0.208)                     & 0.789 (0.198)                     & 0.802 (0.177)                     & 0.818 (0.165)                     & 0.824 (0.148)          \\
\multicolumn{1}{c|}{}                         & \multicolumn{1}{l|}{Uniform selection} & \multicolumn{1}{l}{0.759 (0.218)} & \multicolumn{1}{l}{0.770 (0.185)} & \multicolumn{1}{l}{0.799 (0.171)} & \multicolumn{1}{l}{0.814 (0.163)} & 0.824 (0.148)          \\ \hline
\multicolumn{1}{c|}{\multirow{3}{*}{PED}}     & Ours                     & \textbf{0.646 (0.206)}            & \textbf{0.676 (0.209)}            & \textbf{0.691 (0.215)}            & \textbf{0.704 (0.207)}            & \textbf{0.724 (0.196)} \\
\multicolumn{1}{c|}{}                         & Random selection                       & 0.633 (0.221)                     & 0.651 (0.227)                     & 0.687 (0.217)                     & 0.689 (0.228)                     & 0.724 (0.196)          \\
\multicolumn{1}{c|}{}                         & \multicolumn{1}{l|}{Uniform selection} & \multicolumn{1}{l}{0.620 (0.206)} & \multicolumn{1}{l}{0.659 (0.209)} & \multicolumn{1}{l}{0.673 (0.218)} & \multicolumn{1}{l}{0.676 (0.237)} & 0.724 (0.196)          \\ \hline
\multicolumn{1}{c|}{\multirow{3}{*}{Choroid}} & Ours                     & \textbf{0.875 (0.076)}            & \textbf{0.879 (0.077)}            & \textbf{0.886 (0.066)}            & \textbf{0.891 (0.076)}            & \textbf{0.895 (0.073)} \\
\multicolumn{1}{c|}{}                         & Random selection                       & 0.872 (0.071)                     & 0.876 (0.078)                     & 0.881 (0.080)                     & 0.884 (0.079)                     & 0.895 (0.073)          \\
\multicolumn{1}{c|}{}                         & \multicolumn{1}{l|}{Uniform selection} & \multicolumn{1}{l}{0.872 (0.087)} & \multicolumn{1}{l}{0.875 (0.078)} & \multicolumn{1}{l}{0.880 (0.086)} & \multicolumn{1}{l}{0.890 (0.084)} & 0.895 (0.073)          \\ \hline
\end{tabular}
}
\end{table}
% \usepackage{multirow}
% \usepackage{graphicx}
\indent Without both Phase 1 and 2 (using the Transformer to CNN architecture in Phase 4 directly), The segmentation accuracies are poor for the SRF and PED. After including Phase 1 (the masked self-supervised generative pre-training), their performances significantly improve. We further compare three different training settings used in Phase 1. The model achieves the best accuracy when trained on the ImageNet and fine-tuned on the OCT2017 dataset. The addition of Phase 2 further improves the segmentation performance, which justifies the contribution of the segmentation pre-training.\\
\indent For the Choroid, due to its morphological features do not vary much in different locations of the fundus and for different cases, we can achieve decent segmentation accuracies using the U-Net with a data budget of 10\% as mentioned above. Here we have a similar observation that the benefits of using Phase 1 and 2 pre-training are relatively small.\\
%1) Effect of pre-training phases. According to our process pipeline, Phase 1 and Phase 2 play different roles, respectively. Three ablation settings for without Phase 1 and 2, with Phase 1 and without Phase 2, with Phase 1 and 2 are preformed to compare the segmentation accuracy. We also explore the performance of Phase 1 pre-trained by ImageNet or/and OCT dataset. Table \ref{tab:my-table2} shows the corresponding quantitative evaluation results in SRF, PED, Choroid targets with 10\% and 100\% data budget,respectively. It can be seen that the segmentation accuracy gradually improves as the addition of Phase 1 and Phase 2. 
\indent To verify the effectiveness of Phase 3 in our proposed method, we compare the segmentation accuracies using our selective annotation algorithm with those using random selection and uniform selection (taking B-scans every constant distance from each case) as shown in Table~\ref{tab:my-table2}. For the three types of ROIs, the proposed selection achieves superior performances under all the data budgets we tested ranging from 5\% to 50\%. 

% % Please add the following required packages to your document preamble:
% % \usepackage{multirow}
% % \usepackage{graphicx}
% \begin{table}[h]
% \centering
% \caption{Comparison of the proposed selection with random selection. The format of the data in the table is the mean (standard deviation).}
% \label{tab:my-table2}
% \resizebox{12cm}{!}{%
% \begin{tabular}{cc|ccccc}
% \hline
% \multicolumn{2}{c|}{Setting} & 5\% & 10\% & 25\% & 50\% & 100\% \\ \hline
% \multicolumn{1}{c|}{\multirow{2}{*}{SRF}} & Ours & \textbf{0.776 (0.199)} & \textbf{0.795 (0.186)} & \textbf{0.809 (0.161)} & \textbf{0.821 (0.155)} & \textbf{0.824 (0.148)} \\
% \multicolumn{1}{c|}{} & Random selection & 0.767 (0.208) & 0.789 (0.198) & 0.802 (0.177) & 0.818 (0.165) & 0.824 (0.148) \\ \hline
% \multicolumn{1}{c|}{\multirow{2}{*}{PED}} & Ours & \textbf{0.646 (0.206)} & \textbf{0.676 (0.209)} & \textbf{0.691 (0.215)} & \textbf{0.704 (0.207)} & \textbf{0.724 (0.196)} \\
% \multicolumn{1}{c|}{} & Random selection & 0.633 (0.221) & 0.651 (0.227) & 0.687 (0.217) & 0.689 (0.228) & 0.724 (0.196) \\ \hline
% \multicolumn{1}{c|}{\multirow{2}{*}{Choroid}} & Ours & \textbf{0.875 (0.076)} & \textbf{0.879 (0.077)} & \textbf{0.886 (0.066)} & \textbf{0.891 (0.076)} & \textbf{0.895 (0.073)} \\
% \multicolumn{1}{c|}{} & Random selection & 0.872 (0.071) & 0.876 (0.078) & 0.881 (0.080) & 0.884 (0.079) & 0.895 (0.073) \\ \hline
% \end{tabular}%
% }
% \end{table}

% \usepackage{multirow}
\subsection{Comparison with other potential methods}\label{compsec}
We further compare our proposed method with other deep learning techniques that may contribute to improving annotation efficiency in OCT segmentation, including transfer learning \cite{kermany2018identifying} and two semi-supervised learning methods \cite{luo2022semi,lei2022semi}. We refer to the method developed in \cite{luo2022semi} as semi-supervised learning-1 and the method developed in \cite{lei2022semi} as semi-supervised learning-2. For a fair comparison, in the implementation of the transfer learning, we use the vision transformer \cite{dosovitskiy2020image} trained on the ImageNet and fine-tuned on the OCT2017 dataset as the encoder. We then connect it to the CNN decoder described above for the segmentation capability. We also use the DukeOCT dataset to re-train the transfer learning model. In the implementations of the semi-supervised learning methods, we employ all the unlabeled data used in the pre-training of our method.\\
\begin{table}[!h]
\centering
\caption{Comparison with other methods that may improve the annotation efficiency in OCT segmentation. The data format in the table is the mean (standard deviation).}
\label{tab:my-table3}
\resizebox{\textwidth}{!}{%
\begin{tabular}{cl|cc|cc|cc}
\hline
\multicolumn{2}{c|}{\multirow{2}{*}{Methods}} & \multicolumn{2}{c|}{SRF} & \multicolumn{2}{c|}{PED} & \multicolumn{2}{c}{Choroid} \\ \cline{3-8} 
\multicolumn{2}{c|}{} & 10\% & 100\% & 10\% & 100\% & 10\% & 100\% \\ \hline
\multicolumn{2}{c|}{Transfer learning \cite{kermany2018identifying}} & 0.637 (0.278) & 0.727 (0.239) & 0.381 (0.212) & 0.525 (0.254) & 0.875 (0.087) & 0.890 (0.076) \\
\multicolumn{2}{c|}{Semi-supervised learning-1 \cite{luo2022semi}} & 0.784 (0.220) & 0.796 (0.198) & 0.587 (0.244) & 0.622 (0.255) & 0.871 (0.076) & 0.882 (0.095) \\
\multicolumn{2}{c|}{Semi-supervised learning-2 \cite{lei2022semi}} & 0.737 (0.267) & 0.798 (0.193) & 0.536 (0.229) & 0.672 (0.212) & 0.874 (0.081) & 0.888 (0.084) \\
\multicolumn{2}{c|}{Ours} & \textbf{0.795 (0.186)} & \textbf{0.824 (0.148)} & \textbf{0.676 (0.209)} & \textbf{0.724 (0.196)} & \textbf{0.879 (0.077)} & \textbf{0.895 (0.073)} \\ \hline
\end{tabular}%
}
\end{table}

\begin{figure}[!h]
\centering\includegraphics[width=13cm]{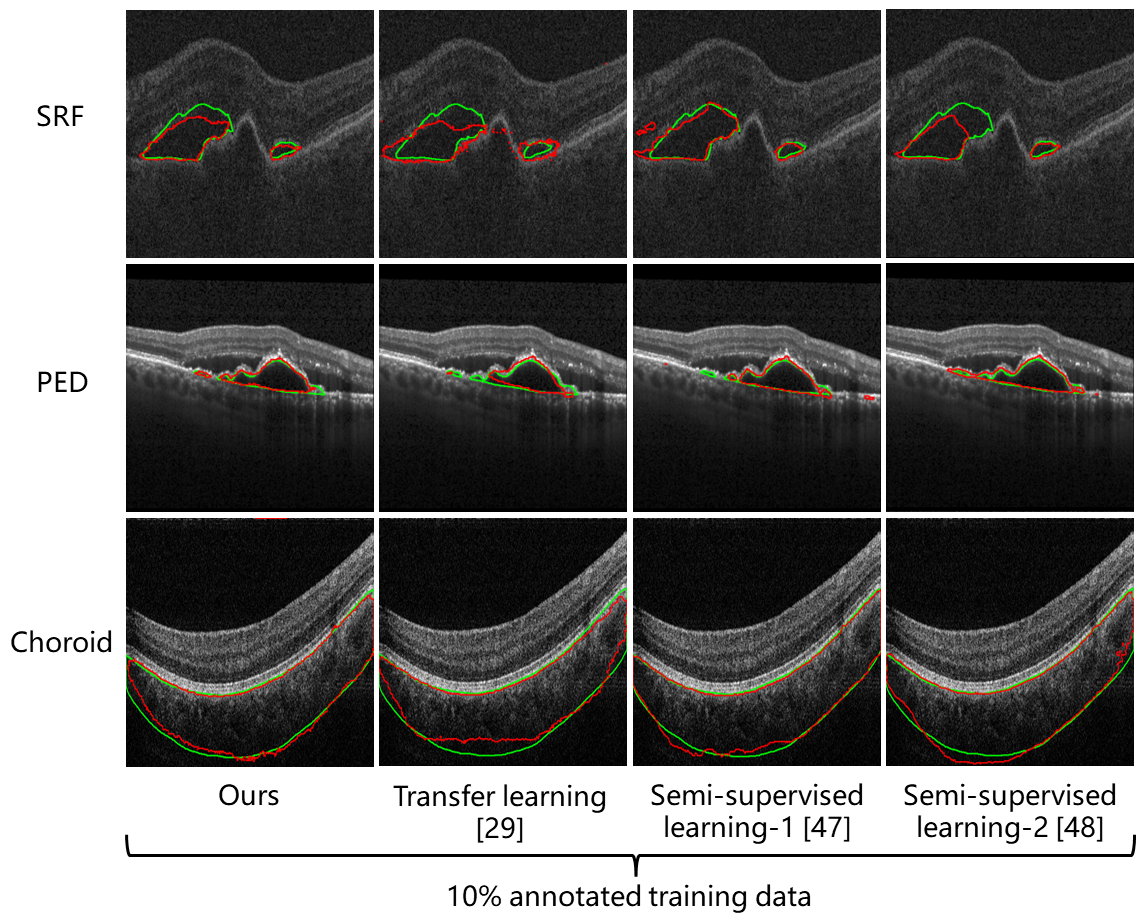}
\caption{Qualitative comparison of our method with other methods that may improve the annotation efficiency in OCT segmentation. The ground truth is labeled in green.}\label{comp}
\end{figure}
\indent Table~\ref{tab:my-table3} gives the quantitative comparison of segmentation accuracy. Our method outperforms other methods on all the ROIs and data budgets. Especially in the segmentation of the SRF and PED at a data budget of 10\%, our method surpasses the transfer learning and the semi-supervised learning methods by a large margin. Figure~\ref{comp} is their qualitative comparison under a data budget of 10\%. Our method achieves better consistency with the ground truth.\\
\begin{table}[h!]
\centering
\caption{Comparison of training time in seconds.}
\label{tab:my-table4}
\resizebox{8cm}{!}{%
\begin{tabular}{cl|cc|cc|cc}
\hline
\multicolumn{2}{c|}{\multirow{2}{*}{Methods}} & \multicolumn{2}{c|}{SRF} & \multicolumn{2}{c|}{PED} & \multicolumn{2}{c}{Choroid} \\ \cline{3-8} 
\multicolumn{2}{c|}{} & 10\% & 100\% & 10\% & 100\% & 10\% & 100\% \\ \hline
\multicolumn{2}{c|}{Transfer learning \cite{kermany2018identifying}} & 232 & 1138 & 178 & 861 & 156 & 432 \\
\multicolumn{2}{c|}{Semi-supervised learning-1 \cite{luo2022semi}} & 205 & \textbf{470} & 397 & 654 & 225 & 363 \\
\multicolumn{2}{c|}{Semi-supervised learning-2 \cite{lei2022semi}} & 1976 & 5399 & 1655 & 3749 & 633 & 852 \\
\multicolumn{2}{c|}{Ours} & \textbf{146} & 615 & \textbf{113} & \textbf{641} & \textbf{125} & \textbf{242} \\ \hline
\end{tabular}%
}
\end{table}
\indent We also compare the training time of different annotation-efficient segmentation methods as shown Table~\ref{tab:my-table4}. We can see that our method is generally faster than other methods. It should be noted that only the re-training time using the selected target data (Phase 4) is listed in this table, because our pre-trainings in Phases 1 and 2 were done once and for all. The pre-trained models are offline and can be adapted to different data sources and ROIs. We also give the time costs of Phases 1 to 3 for reference. The Phase 1 model trained on the OCT2017 dataset spent 55 hours. The Phase 2 model trained on the DukeOCT dataset spent 14 hours. For the selective annotation in Phase 3, it cost 2.5, 2.9, and 1.4 seconds on the SRF, PED, and choroid datasets, respectively.
\section{Discussion}
\indent The experimental results demonstrated above justify the effectiveness of our proposed method in saving the annotation and training time costs of deep-learning-based OCT segmentation, which can be useful in many scenarios, such as surgical navigation and multi-center clinical trials. Methodologically, our method outperforms other methods that may be used to improve annotation efficiency, including transfer learning and semi-supervised learning. The advantages of our method can be summarized below:\\
\indent We leverage the emerging self-supervised generative learning \cite{liu2021self,krishnan2022self}, which has been demonstrated to be effective in transferring to general computer vision tasks \cite{he2022masked,zoph2020rethinking,wei2022masked}, here we use it to address the annotation efficiency problem in OCT segmentation. Instead of directly transferring the trained model via self-supervised generative learning, we introduce an intermediate pre-training step (Phase 2) to further reduce the annotation cost, which has been justified in the ablation studies above. The pre-trained model can be employed to adapt the target data from different manufacturers and imaging protocols, and for different ROIs, which has been verified in the experimental results above. The data we used involve two sources of data (the RETOUCH challenge and a local hospital), three types of ROI (the SRF, PED, Choroid), and four different manufacturers (ZEISS, Heidelberg, Topcon, and a homemade SD OCT system).\\
\indent We collect publicly available OCT datasets and employ them in our training Phase 1 and 2. They help us shrink the hypothesis space \cite{wang2020generalizing}. Semi-supervised learning techniques also use unlabeled data to boost segmentation performance \cite{luo2022semi, lei2022semi}. But they usually require that both labeled and unlabeled data are from the same manufacturer and use the same imaging protocol, which limits their generalization capability. Besides, their training procedures are usually more complicated and time-consuming than the pre-training strategy, which brings inconvenience to end users. These arguments have been justified in the comparison with state-of-the-art semi-supervised learning segmentation methods in Section \ref{compsec}.\\
\indent We design a selective annotation algorithm for avoiding the redundancy and inefficiency that existed among adjacent OCT B-scans from the same case. This algorithm can effectively perform the core-set selection by geometrically approximating the feature space of the entire training set with a small number of samples. Different from active learning techniques \cite{sener2018active} that require the interaction among model training, sample selection, and annotation, our method does not involve any labels and model training, and is therefore more efficient and easy to use. The results show our method can contribute to the improvement of segmentation accuracy.\\
\indent Despite the above advantages of our method, there is still room for improvement in terms of annotation efficiency and training speed, especially the training speed is far from the requirements for applications such as surgical navigation and monitoring. In addition, the robustness of the method still needs to be validated using more external data.

\section{Conclusion}
We have developed a method for improving the annotation efficiency of deep-learning-based OCT segmentation. Compared to the widely-used U-Net model with 100\% training data, our method only requires $\sim10\%$ of the annotated data for achieving the same segmentation accuracy, and it speeds the training up to $\sim3.5$ times. Moreover, our proposed method outperforms other potential strategies that could improve annotation efficiency in OCT segmentation. We think this emphasis on learning efficiency may help improve the intelligence and application penetration of OCT-based technologies.

\begin{backmatter}
\bmsection{Funding}
This work was supported by the National Natural Science Foundation of China under Grant No.~51890892 and 62105198.
\bmsection{Disclosures}
The authors declare no conflicts of interest.
\bmsection{Data availability}
Data underlying the results presented in this paper are available in Ref. \cite{kermany2018identifying,farsiu2014quantitative,bogunovic2019retouch}.
\end{backmatter}

%%%%%%%%%%%%%%%%%%%%%%% References %%%%%%%%%%%%%%%%%%%%%%%%%
%%%%%%%%%% If using BibTeX:
\bibliography{sample}

\end{document}